\documentclass[10pt,journal,compsoc]{IEEEtran}
\usepackage{amsfonts,amsmath,amssymb,amsthm,bm}
\usepackage{booktabs,threeparttable}
\usepackage{graphicx,psfrag,epsf}
\usepackage{enumerate}
\usepackage{url}
\usepackage{algorithm}
\usepackage{algpseudocode}
\usepackage{helvet}
\usepackage[colorlinks=true,pagebackref,linkcolor=magenta]{hyperref}
\usepackage{color} 
\usepackage{lineno}
\usepackage{chngcntr}
\usepackage{thmtools, thm-restate}

\theoremstyle{plain}

\usepackage{adjustbox}
\usepackage{dsfont}
\ifCLASSOPTIONcompsoc
  \usepackage[nocompress]{cite}
\else
  \usepackage{cite}
\fi

\ifCLASSINFOpdf

\else
 
\fi

\newcommand{\blind}{1}

\begin{document}
  \title{\bf One-Hot Graph Encoder Embedding}
  \author{Cencheng Shen, Qizhe Wang, Carey E. Priebe
  \IEEEcompsocitemizethanks{
  \IEEEcompsocthanksitem Cencheng Shen and Qizhe Wang are with the Department of Applied Economics and Statistics, University of Delaware. E-mail: shenc@udel.edu, qizhew@udel.edu 
  \IEEEcompsocthanksitem Carey E.Priebe
is with
the Department of Applied Mathematics and Statistics (AMS), the Center for Imaging Science (CIS), and the Mathematical Institute for Data Science (MINDS), Johns Hopkins University. E-mail: cep@jhu.edu \protect
}
\thanks{This work was supported in part by the Defense Advanced Research Projects Agency under the D3M program administered through contract FA8750-17-2-0112,
the National Science Foundation HDR TRIPODS 1934979, the National Science Foundation DMS-2113099, the University of Delaware Data Science Institute Seed Funding Grant,
and by funding from Microsoft Research. We thank the editor and reviewers for their excellent suggestions to improve the paper. We thank Jonathan Larson and Ha Trinh from Microsoft Research for test running our code. 
}}
\markboth{IEEE Transactions on Pattern Analysis and Machine Intelligence, 2023}%
{Shell \MakeLowercase{\textit{et al.}}: Bare Demo of IEEEtran.cls for Computer Society Journals}

\IEEEtitleabstractindextext{%
\begin{abstract} 
In this paper we propose a lightning fast graph embedding method called one-hot graph encoder embedding. It has a linear computational complexity and the capacity to process billions of edges within minutes on standard PC --- making it an ideal candidate for huge graph processing. 
It is applicable to either adjacency matrix or graph Laplacian, and can be viewed as a transformation of the spectral embedding. Under random graph models, the graph encoder embedding is approximately normally distributed per vertex, and asymptotically converges to its mean. We showcase three applications: vertex classification, vertex clustering, and graph bootstrap. In every case, the graph encoder embedding exhibits unrivalled computational advantages.
\end{abstract}

}

\maketitle
\IEEEdisplaynontitleabstractindextext
\IEEEpeerreviewmaketitle

\IEEEraisesectionheading{\section{Introduction}}

\IEEEPARstart{G}{raph} data arises naturally in modern data collection and captures interactions among objects. Given $n$ vertices and $s$ edges, a graph can be represented by an $n \times n$ adjacency matrix $\mathbf{A}$ where $\mathbf{A}(i,j)$ is the edge weight between $i$th vertex and $j$th vertex. In practice, a graph is typically stored by an $s \times 3$ edgelist $\mathbf{E}$, where the first two columns store the vertex indices of each edge and the last column is the edge weight. Examples include social networks, brain regions, article hyperlinks \cite{GirvanNewman2002, AdamicGlance2005,Newman2006,VogelsteinPark2014, src1}, etc. A graph data has community structure if the vertices can be grouped into different classes based on the edge connectivity \cite{GirvanNewman2002}. In case of supervised learning, some vertices come with ground-truth labels and serve as the training data; while in case of unsupervised learning, the graph data has no known label.   

To better explore and analyze graph data, graph embedding is a very popular approach, which learns a low-dimensional Euclidean representation of each vertex. The spectral embedding method \cite{RoheEtAl2011, SussmanEtAl2012, TangSussmanPriebe2013,SussmanTangPriebe2014,Tang2017,Priebe2019} is a well-studied method in the statistics literature. By using singular value decomposition (SVD) on graph adjacency or graph Laplacian, the resulting vertex embedding asymptotically converges to the latent positions under random dot product graphs \cite{HollandEtAl1983, YoungScheinerman2007}, thus consistent for subsequent inference tasks like hypothesis testing and community detection. Other popular approaches include Deepwalk \cite{perrozi2014}, node2vec \cite{grover2016node2vec,node2vec2021}, graph convolutional network (GCN) \cite{kipf2016semi}, which empirically work well on real graphs. However, existing methods require tuning parameters, are computationally expensive, and do not scale well to big graphs. As modern social networks easily produce billions of edges, a more scalable and elegant solution is direly needed. 

Towards that target, we propose the one-hot graph encoder embedding (GEE) in this paper. The method is straightforward to implement in any programming language, has a linear computational complexity and storage requirement, is applicable to either the adjacency matrix or graph Laplacian, and is capable of processing billions of edges within minutes on a standard PC. Theoretically, the graph encoder embedding enjoys similar properties as the spectral embedding, is approximately normally distributed, and converges to a transformation of the latent positions under random graph models. We showcase three applications: vertex classification, vertex clustering, and graph bootstrap. Comprehensive experiments on synthetic and real graphs are carried out to demonstrate its excellent performance. All proofs and simulation details are in the Appendix. The MATLAB, Python, and R code are made available on Github\footnote{\url{https://github.com/cshen6/GraphEmd}}.

\section{Method}
\label{main}
\subsection*{Graph Encoder Embedding}
\label{sec1}

Algorithm~\ref{alg0} presents the pseudo-code for encoder embedding when all or partial vertex labels are available. The inputs consist of an edgelist $\mathbf{E}$ and a label vector $\mathbf{Y}$ of $K$ classes. We assume the known labels lie in $\{1,\ldots,K\}$ and unknown labels are set to $0$ (or any negative number suffices). The final embedding is denoted by $\mathbf{Z}$, where $\mathbf{Z}_i$ (the $i$th row) is the embedding of the $i$th vertex.

\begin{algorithm}
\caption{Graph Encoder Embedding}
\label{alg0}
\begin{algorithmic}
\Require An edgelist $\mathbf{E} \in \mathbb{R}^{s \times 3}$, and the corresponding class label vector $\mathbf{Y} \in \{0,\ldots,K\}^{n}$.
\Ensure The encoder embedding $\mathbf{Z} \in \mathbb{R}^{n \times K}$, and the transformation matrix $\mathbf{W} \in \mathbb{R}^{n \times K}$.
\Function{GEE}{$\mathbf{E},\mathbf{Y}$}
\State $\mathbf{W}=\operatorname{zeros}(n,K)$; \Comment{initialize the matrix}
\State $\mathbf{Z}=\operatorname{zeros}(n,K)$; 
\For{$k=1,\ldots,K$}
\State $ind=\operatorname{find}(\mathbf{Y}=k)$; \Comment{find indices of class $k$}
\State $n_k= \operatorname{sum}(ind)$; 
\State $\mathbf{W}(ind,k) = \frac{1}{n_{k}}$;
\EndFor
\For{$i=1,\ldots,s$}
\State $\mathbf{Z}(\mathbf{E}(i,1),\mathbf{Y}(\mathbf{E}(i,2)))=\mathbf{Z}(\mathbf{E}(i,1),\mathbf{Y}(\mathbf{E}(i,2)))+\mathbf{W}(\mathbf{E}(i,2),\mathbf{Y}(\mathbf{E}(i,2)))*\mathbf{E}(i,3)$; 
\State $\mathbf{Z}(\mathbf{E}(i,2),\mathbf{Y}(\mathbf{E}(i,1)))=\mathbf{Z}(\mathbf{E}(i,2),\mathbf{Y}(\mathbf{E}(i,1)))+\mathbf{W}(\mathbf{E}(i,1),\mathbf{Y}(\mathbf{E}(i,1)))*\mathbf{E}(i,3)$; 
\EndFor
\EndFunction
\end{algorithmic}
\end{algorithm}

The algorithm is applicable to any graph, including directed or weighted graphs. It is also applicable to the graph Laplacian: given any edgelist, one can compute the degree coefficient for each vertex, then replace the edge weight by the degree-normalized weight. This can be achieved via iterating through the edgelist just twice (not shown in Algorithm~\ref{alg0} but implemented in our codebase). 

Since $n_k$ represents the number of vertices in each class, the matrix $\mathbf{W}$ equals the one-hot encoding of the label vector then column-normalized by $n_k$. In matrix notation, the encoder embedding can be succinctly expressed by $\mathbf{Z}=\mathbf{A}\mathbf{W}$, or $\mathbf{Z}=\mathbf{D}^{-0.5} \mathbf{A} \mathbf{D}^{-0.5}\mathbf{W}$ for graph Laplacian ($\mathbf{D}$ is the $n \times n$ diagonal matrix of degrees). 

In the one-hot graph encoder embedding, each class label of the graph vertex is assigned its own variable in the final embedding. We shall call the adjacency version as the adjacency encoder embedding (AEE), and the Laplacian version as the Laplacian encoder embedding (LEE). They may be viewed as a transformation of the adjacency / Laplacian spectral embedding (ASE / LSE), each with its unique property as summarized in \cite{Priebe2019}. Note that Algorithm~\ref{alg0} assumes partial known labels and is a natural set-up for vertex classification, which is evaluated in-depth in Section~\ref{main2}. The unsupervised GEE (no known label) is presented in Algorithm~\ref{alg3} and evaluated in Section~\ref{main3}.

\subsection*{Computational Advantages}

\begin{figure}
\centering
\includegraphics[width=0.9\linewidth,trim={0cm 0cm 0cm 1cm},clip]{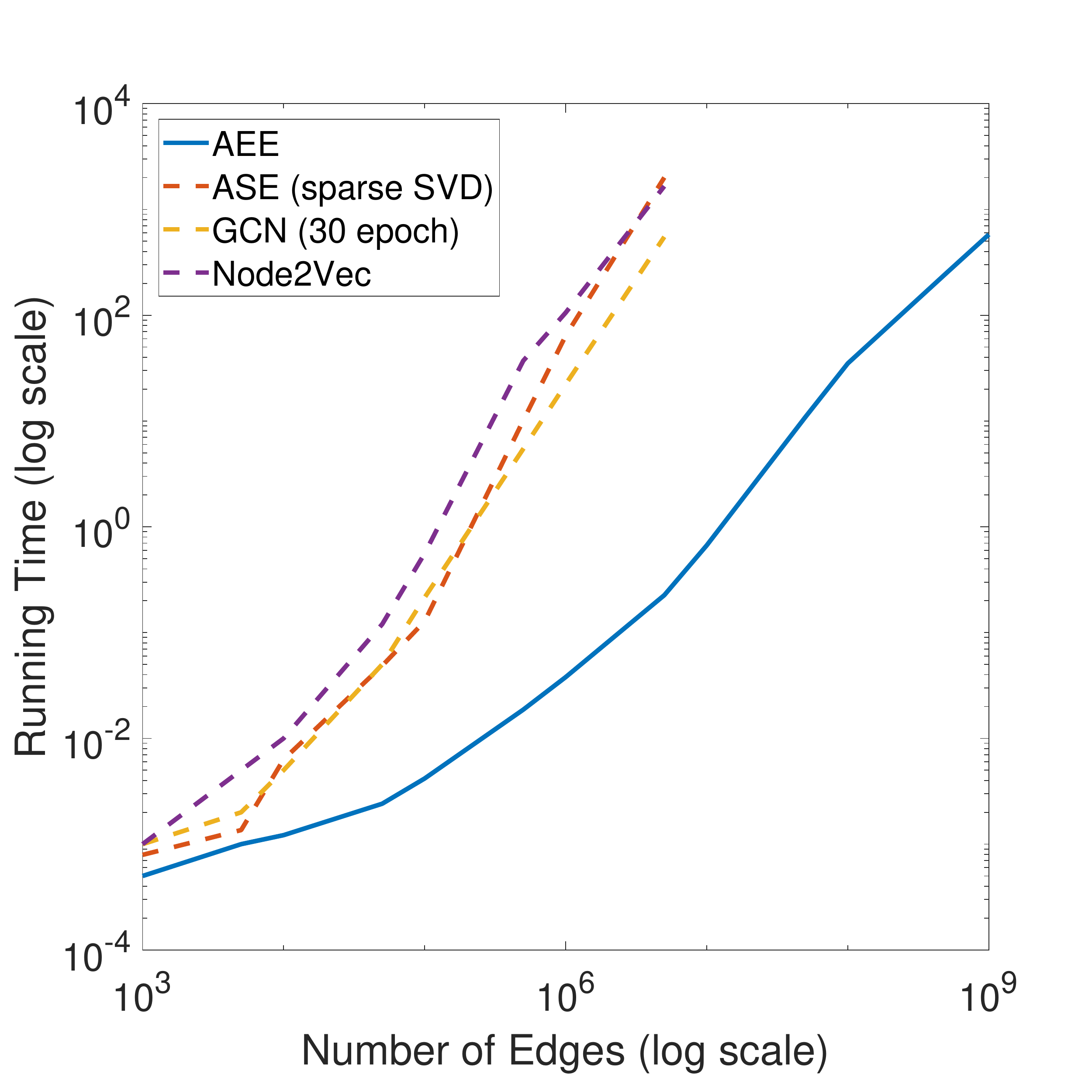}
\caption{We report the average running time of graph encoder embedding using $50$ Monte Carlo replicates, on a random graph with $K=10$, average degree $100$, and increasing graph size. The number of edges increases from one thousand to one billion. At $1$ billion edges with $10$ million vertices, the encoder embedding only requires $20$GB memory and finishes in $10$ minutes. All other methods exceed maximum memory capacity at $10$ million edges. More details on the methods compared can be found in Section~\ref{main2}.}
\label{fig4}
\end{figure}

Algorithm~\ref{alg0} has a time complexity and storage requirement of $O(nK+s)$, thus is linear with respect to the number of vertices and number of edges. Because it iterates the input data only once with a few operations, it is extremely efficient in any programming language. The running time advantage is demonstrated in Figure~\ref{fig4}. On a standard PC with 12-core CPU and 64GB memory and MATLAB 2022a, it takes a mere 6 seconds to process $10$ million edges, one minute for $100$ million edges with $1$ million vertices, and $10$ minutes for $1$ billion edges with $10$ million vertices. In comparison, other methods are order of magnitude slower and cannot handle more than $10$ million edges on the same PC. 

\section{Theorems}

To better understand graph encoder embedding, we first review three popular random graph models, then present the asymptotic properties under each model. Throughout this section, we assume $n$ is the number of vertices with known labels; and when $n \rightarrow \infty$, so is $n_k \rightarrow \infty$ for each $k \in \{1,\ldots,K\}$.

\subsection*{Stochastic Block Model (SBM)}

SBM is arguably the most fundamental community-based random graph model \cite{HollandEtAl1983, SnijdersNowicki1997, KarrerNewman2011, Gao2017}. Each vertex $i$ is associated with a class label $Y_i \in \{1,\ldots, K\}$. The class label may be fixed a-priori, or generated by a categorical distribution with prior probability $\{\pi_k \in (0,1) \mbox{ with }  \sum_{k=1}^{K} \pi_k=1\}$. Then a block probability matrix $\mathbf{B}=[\mathbf{B}(k,l)] \in [0,1]^{K \times K}$ specifies the edge probability between a vertex from class $k$ and a vertex from class $l$: for any $i<j$,
\begin{align*}
\mathbf{A}(i,j) &\stackrel{i.i.d.}{\sim} \operatorname{Bernoulli}(\mathbf{B}(Y_i, Y_j)), \\
\mathbf{A}(i,i)&=0, \ \ \mathbf{A}(j,i) = \mathbf{A}(i,j).
\end{align*}

\subsection*{Degree-Corrected Stochastic Block Model (DC-SBM)}

The DC-SBM graph is a generalization of SBM to better model the sparsity of real graphs \cite{ZhaoLevinaZhu2012}. Everything else being the same as SBM, each vertex $i$ has an additional degree parameter $\theta_i$, and the adjacency matrix is generated by
\begin{align*}
\mathbf{A}(i,j) \sim \operatorname{Bernoulli}(\theta_i \theta_j \mathbf{B}(Y_i, Y_j)).
\end{align*}
The degree parameters typically require certain constraint to ensure a valid probability. In this paper we simply assume they are non-trivial and bounded, i.e., $\theta_i \stackrel{i.i.d.}{\sim} F_{\theta} \in (0,M]$, which is a very general assumption. 

\subsection*{Random Dot Product Graph (RDPG)}

Another random graph model is RDPG \cite{YoungScheinerman2007}. Under RDPG, each vertex $i$ is associated with a latent position vector $X_i \stackrel{i.i.d.}{\sim} F_X \in [0,1]^{p}$. $F_X$ is constrained such that $X_i^{T} X_j \in (0,1]$, i.e., the inner product shall be a valid  probability. Then the adjacency matrix is generated by
\begin{align*}
\mathbf{A}(i,j) &\sim \operatorname{Bernoulli}(X_i^{T} X_j).
\end{align*}
To generate communities under RDPG, it suffices to use a K-component mixture distribution, i.e., let $(X_i, Y_i) \stackrel{i.i.d.}{\sim} F_{XY}$ be a distribution on $\mathbb{R}^{p} \times [K]$.

\subsection*{Asymptotic Normality}

Under these random graph models, we prove the central limit theorem for the graph encoder embedding. Namely, the vertex embedding is asymptotically normally distributed per vertex. Since the mean and covariance differ under each model, we introduce some additional notations:
\begin{itemize}
\item Denote $\vec{n}=[n_1,n_2,\cdots,n_k] \in \mathbb{R}^{K}$, and $Diag(\cdot)$ as the diagonal matrix of a vector.
\item Under SBM with block matrix $\mathbf{B}$, define $\Sigma_{\mathbf{B}_y}$ as the $K \times K$ diagonal matrix with 
\begin{align*}
&\Sigma_{\mathbf{B}_y}(k,k)=\mathbf{B}(y,k)(1-\mathbf{B}(y,k)) \in[0,\frac{1}{4}].
\end{align*}
\item Under DC-SBM with $\{\theta_j \stackrel{i.i.d.}{\sim} F_{\theta}\}$, for any $t$th moment we define:
\begin{align*}
& \bar{\theta}_{k}^{(t)}=E(\theta_j^t | Y_j=k), \\
& \bar{\Theta}^{(t)}=[\bar{\theta}_{(1)}^{(t)},\bar{\theta}_{(2)}^{(t)},\cdots,\bar{\theta}_{(K)}^{(t)}]\in \mathbb{R}^{K}.
\end{align*}
\item Under RDPG where $(X, Y) \sim F_{XY} \in \mathbb{R}^{p} \times [K]$ is the latent distribution, define
\begin{align*}
\bar{\lambda}_k^{(t)}(x_i)&=E^{t}(X^{T}x_i| Y=k),\\
\bar{\lambda}^{(t)}_{x_i}&=[\bar{\lambda}_1^{(t)}(x_i), \bar{\lambda}_2^{(t)}(x_i), \cdots, \bar{\lambda}_K^{(t)}(x_i)] \in \mathbb{R}^{K}
\end{align*}
for any fixed vector $x_i \in \mathbb{R}^{p}$.
\end{itemize}

\begin{restatable}{theorem}{clt}
\label{thmclt}
The graph encoder embedding is asymptotically normally distributed under SBM, DC-SBM, or RDPG. Specifically, as $n$ increases, for a given $i$th vertex of class $y$ it holds that
\begin{align*}
Diag(\vec{n})^{0.5} \cdot (\mathbf{Z}_{i} - \mu)  \stackrel{d}{\rightarrow}  \mathcal{N}(0,\Sigma).
\end{align*}
The expectation and covariance are:
\begin{itemize}
\item under SBM, 
$\mu=\mathbf{B}(y,:)$ and $\Sigma=\Sigma_{\mathbf{B}_y}$;
\item under DC-SBM, 
$\mu=\theta_i \mathbf{B}(y,:) \odot \bar{\Theta}^{(1)}$ and $\Sigma=\theta_i^2 Diag(\bar{\Theta}^{(2)}) \cdot \Sigma_{\mathbf{B}_y}$;
\item under RDPG,  
$\mu=\bar{\lambda}^{(1)}_{x_i}$ and $\Sigma=Diag(\bar{\lambda}^{(1)}_{x_i} - \bar{\lambda}^{(2)}_{x_i})$.
\end{itemize}
\end{restatable}



\subsection*{Asymptotic Convergence}

The law of large numbers immediately follows. Namely, as the number of vertices increase, the graph encoder embedding converges to the mean.

\begin{restatable}{corollary}{one}
Using the same notation as in Theorem~\ref{thmclt}. It always holds that
\label{thm1}
\begin{align*}
& \|\mathbf{Z}_i -\mu\|_2 \stackrel{n \rightarrow \infty}{\rightarrow} 0.
\end{align*}
\end{restatable}


As SBM, DC-SBM, and RDPG are the most common graph models, in this paper we choose to express the mean via model parameters. Alternatively, the mean can be expressed more generally by conditional expectations, i.e., for each dimension it holds that $\mathbf{Z}_{i}[k] \rightarrow E(\mathbf{A}_{ij} |Y_{j} = k)$, which estimates the probability of vertex $i$ being adjacent to a random vertex from class $k$. 

While the spectral embedding estimates the block probability or latent variable up-to rotation \cite{RoheEtAl2011, SussmanEtAl2012}, the encoder embedding is more informative and interpretable due to the elimination of the rotational non-identifiability. See Figure~\ref{fig1} - \ref{fig5} for numerical examples. Finally, the asymptotic normality and asymptotic convergence also hold for weighted graphs, which is discussed in the proof section.

\section{Embedding Visualization}
\subsection*{Simulated Graphs}
\label{exp1}

\begin{figure}
\centering
\includegraphics[width=1.0\linewidth,trim={2cm 0cm 2cm -0.2cm},clip]{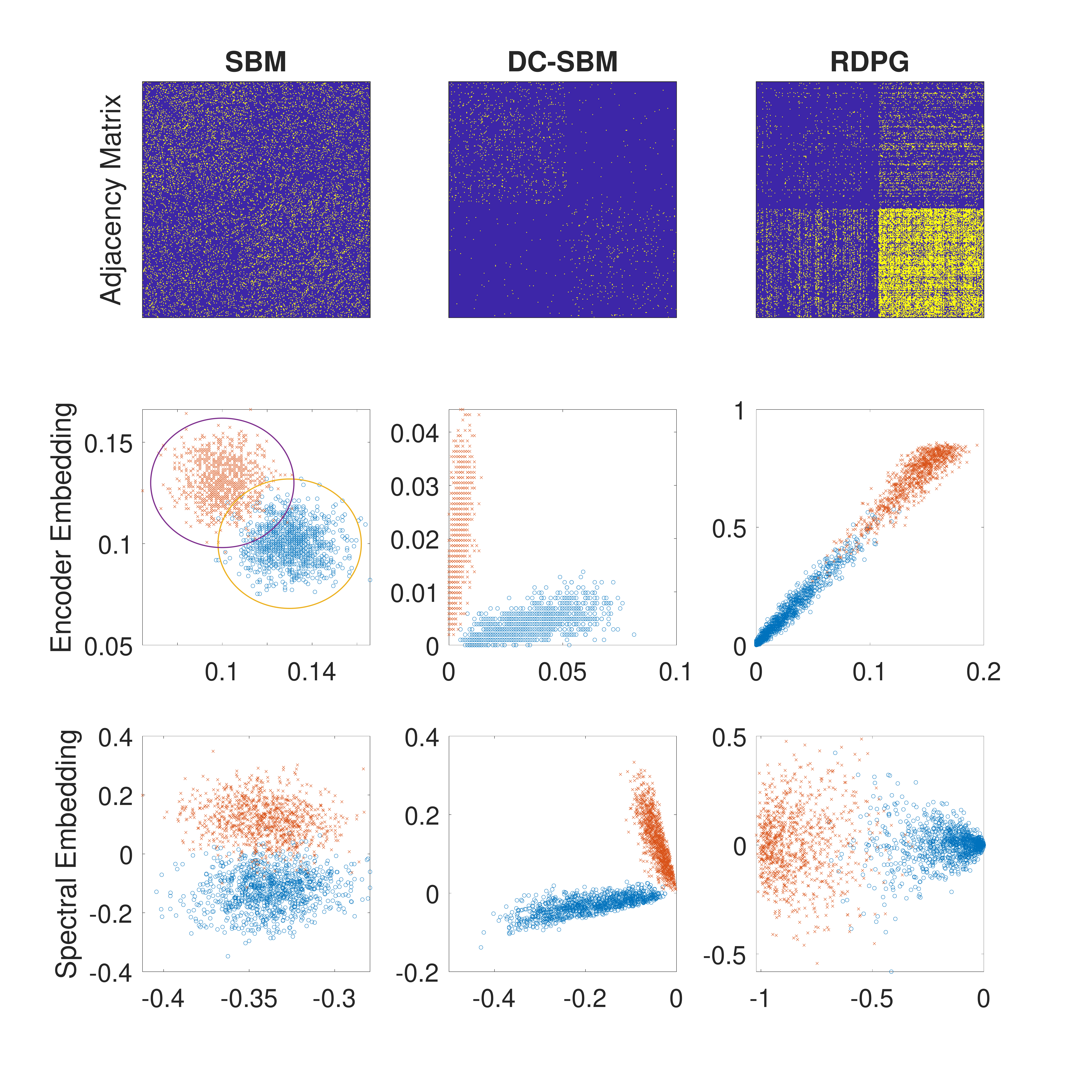}
\caption{Visualizing the vertex embedding: the top row is the graph adjacency heatmap (the index are ordered based on class labels), the middle row is the graph encoder embedding, and the bottom row is the adjacency spectral embedding at $d=2$. Each graph is generated by SBM, DC-SBM, and RDPG from left column to right column at $n=2000$, with parameter details presented in the Appendix. In each panel, the red dots denote the vertex embedding of class $1$, and blue dots denote the vertex embedding of class $2$. }
\label{fig1}
\end{figure}

Figure~\ref{fig1} compares the graph encoder embedding to the spectral embedding under SBM, DC-SBM, and RDPG graphs at $K=2$. While both methods exhibit clear community separation, the encoder embedding provides better estimation for the model parameters. For example, under the SBM graph, the encoder embedding clearly estimates the block probability vectors $(0.13,0.1)$ and $(0.1,0.13)$ and appears normally distributed within each class; and under the DC-SBM graph, the encoder embedding lies along the block probability vectors multiplied by the degree of each vertex. A normality visualization for the same simulations are provided in the Appendix. 

\subsection*{Real Graphs}
\begin{figure}
\centering
\includegraphics[width=1.0\linewidth,trim={1cm 0cm 2cm -0.2cm},clip]{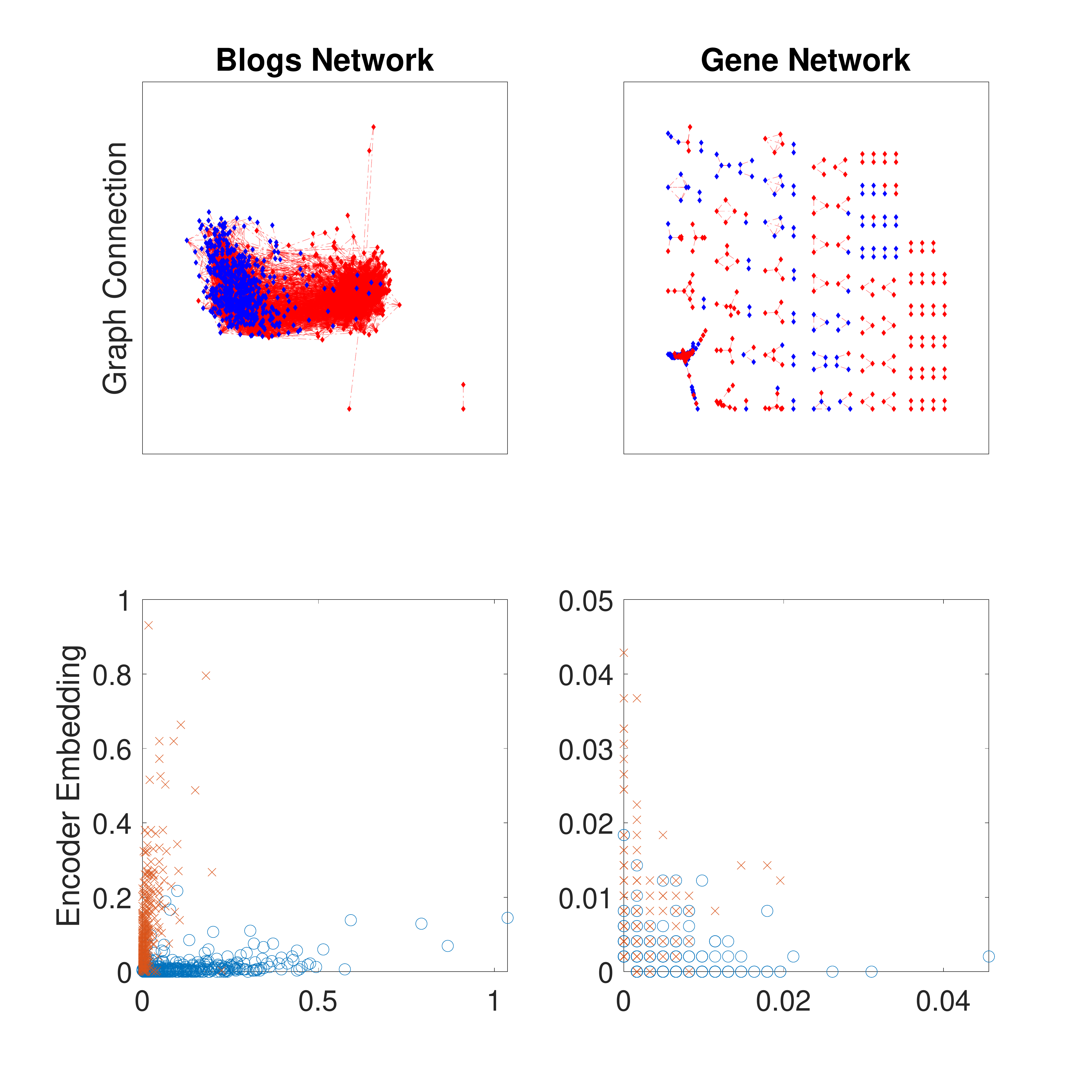}
\caption{Visualizing the vertex embedding for the Political Blogs and Gene Network: the top row plots the graph connectivity via MATLAB graph plot function, and the bottom row is the graph encoder embedding. Red denotes class 1 vertices and blue denotes class 2 vertices. }
\label{fig5}
\end{figure}

Figure~\ref{fig5} illustrates graph encoder embedding for the Political Blogs \cite{AdamicGlance2005} ($1490$ vertices with $2$ classes) and the Gene Network \cite{nr} ($1103$ vertices with $2$ classes). Both graphs are sparse. The average degree is $22.4$ for the Political Blogs and $1.5$ for the Gene Network. We observe that the vertex embedding appears similar to DC-SBM, which lies along a line for each class. Within-class vertices are better connected than between-class vertices, and different communities are well-separated except a few outliers. 

\section{Vertex Classification}
\label{main2}

An immediate and important use case herein is vertex classification. The vertex embedding with known class labels are the training data (labels with class $1$ to $K$), while the vertex embedding with unknown labels are the testing data (labels set to $0$ in Algorithm~\ref{alg0}). We consider five graph embedding methods: adjacency encoder embedding (AEE), Laplacian encoder embedding (LEE), adjacency spectral embedding (ASE), Laplacian spectral embedding (LSE), and node2vec. For ASE and LSE we used the sparse SVD (the fastest SVD implementation in MATLAB) with 20 eigenvalues, then report the best accuracy and the running time among $d=1,\ldots,20$. For node2vec, we use the fastest available PecanPy implementation \cite{node2vec2021} with all default parameters and window size $2$. For every embedding, we use linear discriminant analysis (LDA) and 5-nearest-neighbor (5NN) as the follow-on classifiers. Other classifier like logistic regression, random forest, and neural network can also be used. We observe similar accuracy regardless of the classifiers, implying that the learning task largely depends on the embedding method.

\subsection*{Classification Evaluation on Synthetic Data}

\begin{figure}
\centering
\includegraphics[width=1.0\linewidth,trim={2cm 0cm 2cm -0.2cm},clip]{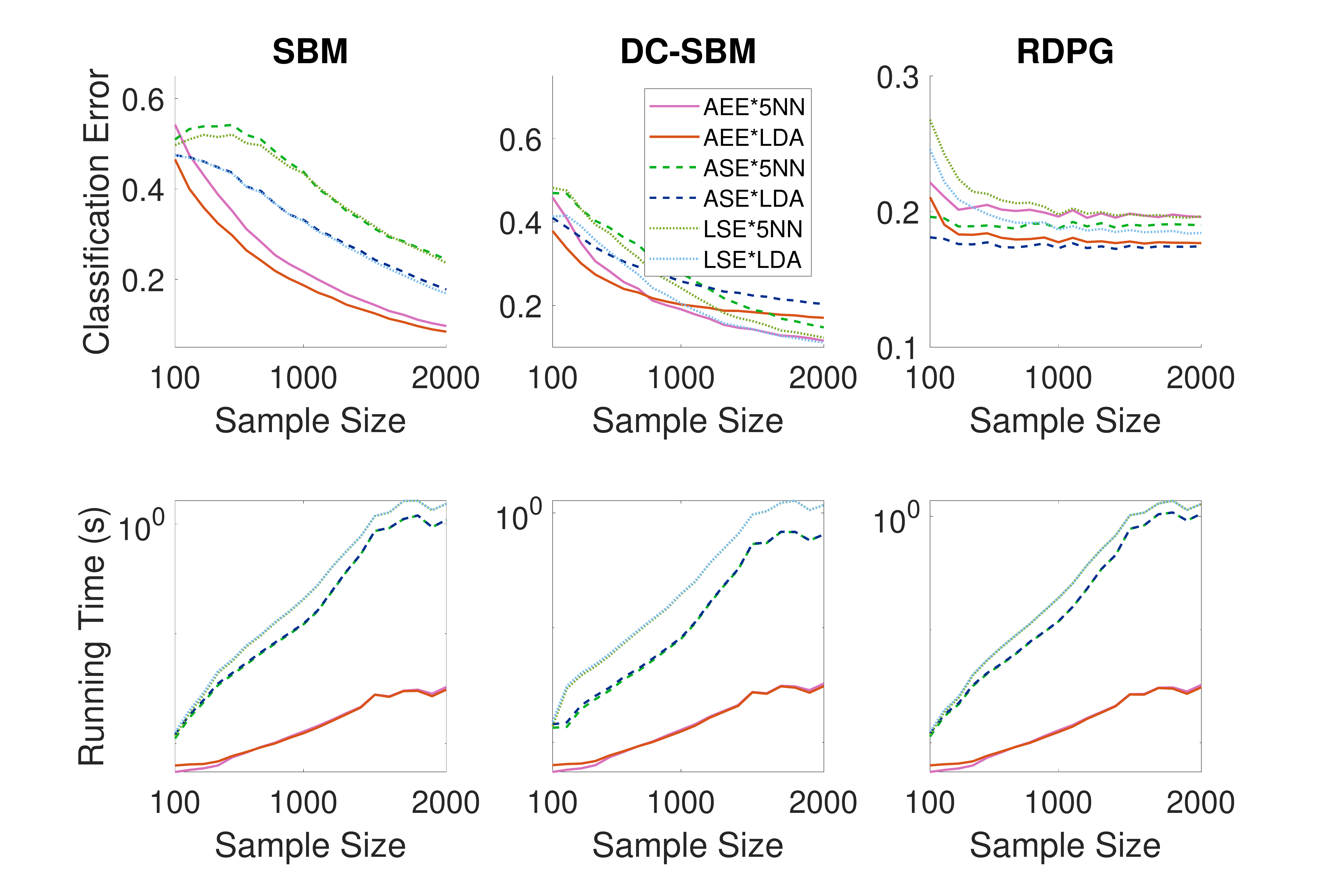}
\caption{Comparing the classification error (top row) and running time (bottom row in log scale) for SBM, DC-SBM, and RDPG graph with increasing $n$. Parameter details can be found in the Appendix.}
\label{fig0}
\end{figure}

Figure~\ref{fig0} shows the average 10-fold classification error and average running time under simulated SBM, DC-SBM, and RDPG graphs ($K=3$). For better clarity, only AEE, ASE, and LSE are included in the figure since they are the best performers on synthetic data. The standard deviation for the classification error is about $2\%$ for each method, while the standard deviation for the running time is at most $10\%$. As the number of vertices increases, every method has better classification error at the cost of more running time. The encoder embedding has the lowest classification error under SBM, is among the lowest under DC-SBM and RDPG, and has the best running time. 

\subsection*{Classification Evaluation on Real Graphs}
We downloaded a variety of public real graphs with labels, including three graphs from network repository\footnote{\url{https://networkrepository.com/index.php}} \cite{nr}: Cora Citations ($2708$ vertices, $5429$ edges, $7$ classes), Gene Network ($1103$ vertices, $1672$ edges, $2$ classes), Industry Partnerships ($219$ vertices, $630$ edges, $3$ classes); and three more graphs from Stanford network data\footnote{\url{https://snap.stanford.edu/}}: EU Email Network \cite{Leskovec2017} ($1005$ vertices, $25571$ edges, $42$ classes), LastFM Asia Social Network \cite{lastfm} ($7624$ vertices, $27806$ edges, $17$ classes), and Political Blogs \cite{AdamicGlance2005} ($1490$ vertices, $33433$ edges, $2$ classes). 

For each data and each method, we carried out 10-fold validation and report the average classification error and running time in Table~\ref{table1}. For ease of presentation, we report the lower error between 5NN and LDA classifiers for each embedding. Comparing to the corresponding spectral embedding or node2vec, the encoder embedding achieves similar or better performance with trivial running time. Node2vec also performs well on real data but takes significantly longer.

\begin{table}
\renewcommand{\arraystretch}{1.3}
\centering
{\begin{tabular}{c||c|c|c|c|c|c}
\hline
  \multicolumn{7}{c}{Classification Error} \\\cline{2-5}
 \hline
 & AEE & LEE & ASE & LSE & N2v & *\\
\hline
Cora & 16.3\%  & \textbf{15.5\%} & 31.0\% &33.1\% & 16.3\% & 69.8\%\\
\hline
Email & 30.6\%  & 28.3\% &30.8\% &39.5\% & \textbf{26.1\%} & 89.2\%\\
\hline
Gene & 17.1\% & \textbf{16.5}\%   &27.2\%  &36.2\% & 21.9\% & 44.4\%\\
\hline
Industry & \textbf{29.7\%} & 30.7\% & 38.8\% &39.2\% & 32.9\% &39.3\%\\
\hline
LastFM & 15.5\%  & 15.0\%  &20.1\% &16.5\% & \textbf{14.5\%} &79.4\% \\
\hline
PolBlog & 4.9\%  & 5.0\% &5.5\% &\textbf{4.0\%} & 4.5\% & 48.0\%\\
\hline
\multicolumn{7}{c}{Running Time (seconds)} \\\cline{2-5}
 \hline
 & AEE & LEE & ASE & LSE & N2v\\
\hline
Cora & \textbf{0.01}  & \textbf{0.01} &1.55 &1.60 & 2.1 \\
\hline
Email & \textbf{0.02} & 0.03 & 0.12 & 0.15 & 1.2 \\
\hline
Gene & \textbf{0.01} & \textbf{0.01} & 0.15 & 0.18 & 0.80 \\
\hline
Industry & \textbf{0.01} & \textbf{0.01}  & 0.02  & 0.02 & 0.25 \\
\hline
LastFM & \textbf{0.02} & 0.03 & 13.0 & 15.3 & 9.2 \\
\hline
PolBlog & \textbf{0.01} & 0.02 &0.27 &0.28 & 1.2 \\
\hline
\end{tabular}
\caption{Comparing the embedding performance on real graphs. For each graph, the lowest classification error and running time are highlighted in bold. N2v stands for node2vec, and the last column shows the chance error. Note that the running time only includes the embedding step. }
\label{table1}
}
\end{table}

\section{No Label and Vertex Clustering}
\label{main3}

Many graph data are collected without ground-truth vertex labels. Therefore, we also design an unsupervised graph encoder embedding in Algorithm~\ref{alg3}. Starting with random label initialization, we utilize Algorithm~\ref{alg0} and k-means clustering to iteratively refine the vertex embedding and label assignments. The algorithm stops when the labels no longer change or the maximum iteration limit is reached.

The running time is $O(M(n K^2+ s))$, which is still linear with respect to the number of edges and the number of vertices. In our experiments we set the maximum iteration limits to $r=30$, which always achieve satisfactory performance. 

Note that spectral embedding and node2vec are unsupervised in nature (though they do not utilize labels even when available). The clustering performance is measured by the adjusted rand index (ARI) between the clustering results and ground-truth labels. ARI lies in $(-\infty,1]$, with larger positive number implying better matchedness and $1$ for perfect match \cite{Rand1971}.

\begin{algorithm}
\caption{Graph Encoder Embedding Without Label}
\label{alg3}
\begin{algorithmic}
\Require An edgelist $\mathbf{E}$, number of clusters $K$, and iteration limit $r$.
\Ensure The encoder embedding $\mathbf{Z} \in \mathbb{R}^{n \times K}$ for all vertices, and the estimated class label $\mathbf{Y} \in \{1,\ldots,K\}^{n}$.
\Function{GEE Unsup}{$\mathbf{E}, K, M$}
\State $\mathbf{Y}_{new}=\operatorname{random}(K,n)$; \Comment{randomize a label vector}
\For{i=1,\ldots,M}
\State $\mathbf{Z}=\operatorname{GEE}(\mathbf{E},\mathbf{Y}_{new})$; 
\State $\mathbf{Y}=\operatorname{kmeans}(\mathbf{Z},K)$;
\If{$\operatorname{ARI}(\mathbf{Y}_{new},\mathbf{Y})==1$}
\State Stop;
\Else
\State $\mathbf{Y}_{new} = \mathbf{Y}$;
\EndIf
\EndFor
\EndFunction
\end{algorithmic}
\end{algorithm}

\begin{figure}
\centering
\includegraphics[width=1.0\linewidth,trim={1cm 0cm 1.8cm -0.2cm},clip]{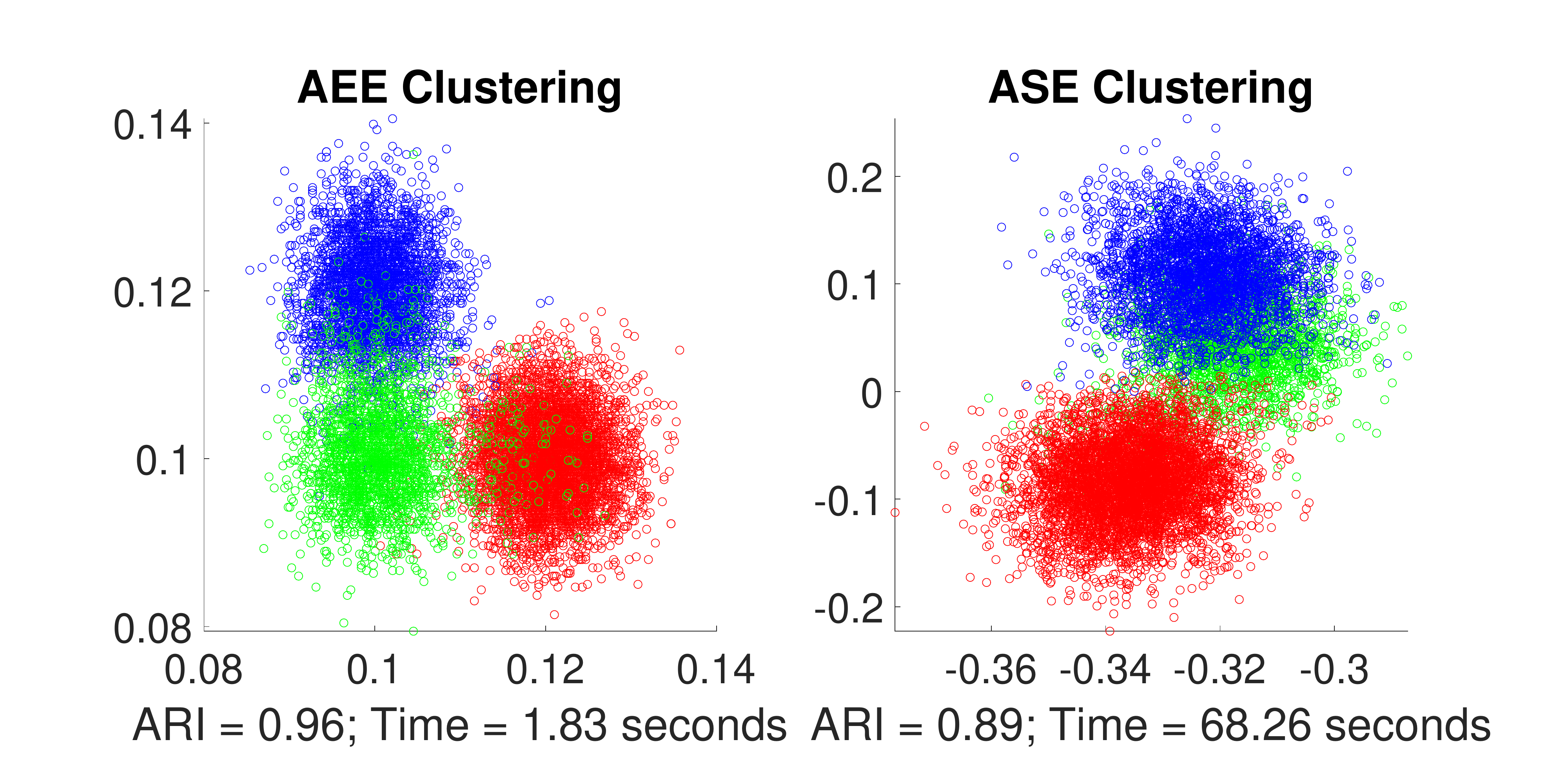}
\centering
\includegraphics[width=1.0\linewidth,trim={1cm 0cm 1.8cm -0.2cm},clip]{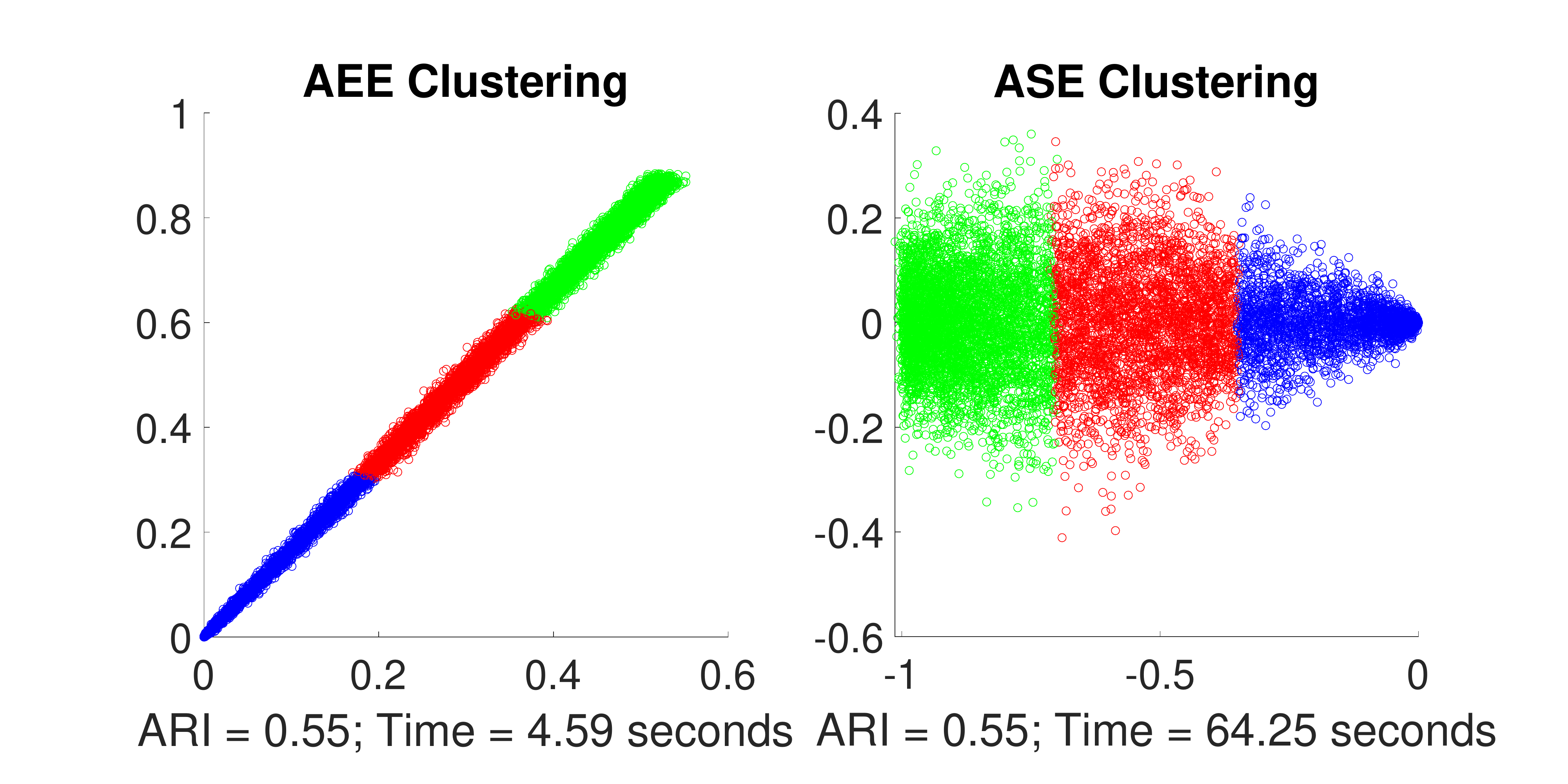}
\caption{The top row visualizes unsupervised AEE and ASE for an SBM graph, while the bottom row compares AEE and ASE for a RDPG graph. Those graphs are generated by the same three-class SBM and RDPG in Figure~\ref{fig0} at $n=10000$. Blue, red, and green dots denote vertices of different classes. Note that the embedding dimension is $3$ while we visualized the embedding of the first two dimensions.}
\label{fig6}
\end{figure}

As long the graph is not too small, Algorithm~\ref{alg3} performs well throughout our experiments. Figure~\ref{fig6} provides an illustration of the clustering performance under 3-class SBM and RDPG graphs. The adjacency encoder embedding yields excellent ARI, which is similar to ASE clustering but much faster. The advantage is consistent throughout the synthetic and real graphs. Table~\ref{table2} presents the clustering results for all the real data in Table~\ref{table1}. Comparing to Table~\ref{table1}, the unsupervised algorithm typically takes $2-10$ times longer than the with-label version. It is still vastly superior than other methods in the running time, while maintaining excellent ARI. The only exception is the Gene graph, which is too sparse for any clustering method. 

\begin{table}[htb]
\renewcommand{\arraystretch}{1.3}
\centering
{\begin{tabular}{c||c|c|c|c|c}
\hline
  \multicolumn{6}{c}{Clustering ARI} \\\cline{2-5}
 \hline
 & AEE & LEE & ASE & LSE & N2v \\
\hline
Cora & 0.12  & 0.07 & 0.08  & 0.01 & \textbf{0.24}\\
\hline
Email & \textbf{0.40}  & 0.39 &0.11 &0.21 & 0.34\\
\hline
Gene & 0.01 & 0.01 & 0.01 & 0.01 & 0.00 \\
\hline
Industry & \textbf{0.13} & 0.03 & 0.01 & 0.02 & \textbf{0.13}\\
\hline
LastFM & 0.34  & 0.19  & 0.03 & \textbf{0.47} & 0.43 \\
\hline
PolBlog & \textbf{0.80} & 0.58 & 0.07 & \textbf{0.80} & \textbf{0.80}\\
\hline
\multicolumn{6}{c}{Running Time (seconds)} \\\cline{2-5}
 \hline
 & AEE & LEE & ASE & LSE & N2v\\
\hline
Cora & \textbf{0.11}  & 0.12 &1.6 &1.7 &2.2\\
\hline
Email & 0.18 & 0.28 &\textbf{0.13} & 0.20 & 1.3\\
\hline
Gene & \textbf{0.03} & \textbf{0.03} & 0.17 & 0.20 & 0.90 \\
\hline
Industry & 0.02 & 0.02  & 0.02  & 0.02 & 0.40\\
\hline
LastFM & \textbf{0.35} & 0.39 & 13.6 & 15.5 & 9.5\\
\hline
PolBlog & \textbf{0.05} & 0.07 & 0.27 &0.29 & 1.4 \\
\hline
\end{tabular}
\caption{K-means clustering results for each embedding method. For each graph, the highest ARI and lowest running time is highlighted in bold. The running time includes both embedding and k-means clustering. }
\label{table2}
}
\end{table}

\section{Graph Bootstrap}
Bootstrap is a popular statistical method for resampling Euclidean data \cite{EfronTibshiraniBook}, and there has been some investigations on graph bootstrap \cite{Bickel2015, Shalizi2022}. A naive graph bootstrap procedure can be carried out as follows: simply resample the vertex index with replacement, then re-index both the row and column of the adjacency matrix.

Since the graph encoder embedding offers a good estimate of the block probability, it also provides an elegant graph bootstrap solution as detailed in Algorithm~\ref{alg2}. Given a graph adjacency and a label vector, we compute the encoder embedding and carry out standard bootstrap on the embedding, then use Bernoulli distribution to form the resampled adjacency matrix. We validate the procedure via a two-sample distance-correlation test \cite{SzekelyRizzoBakirov2007,exact1} between the original and bootstrap graphs via the encoder embedding (testing using graph embedding is asymptotically valid upon mild model assumptions \cite{Tang2017, graph1}). A large p-value suggests that the resampled graph has the same distribution as the original graph, while a small p-value (say less than $0.05$) implies the resampled graph is significantly different in distribution and thus breaking the intention of bootstrap. 

\begin{algorithm}
\caption{Encoder Embedding for Graph Bootstrap}
\label{alg2}
\begin{algorithmic}
\Require $\mathbf{A} \in \mathbb{R}^{n \times n}$, $\mathbf{Y} \in \{1,\ldots,K\}^{n}$, and resampling size $n_2$.
\Ensure Resampled adjacency matrix $\mathbf{A}_{2} \in \mathbb{R}^{n_2 \times n_2}$, corresponding label $\mathbf{Y}_{2} \in \{1,\ldots,K\}^{n_2}$, and a two-sample test p-value $pval$.
\Function{GEEBootstrap}{$\mathbf{A},\mathbf{Y},n_2$}
\State $[\mathbf{Z},\mathbf{W}]=\operatorname{GEE}(\mathbf{A},\mathbf{Y})$;
\State $ind=\operatorname{bootstrap}(n,n_2)$; \Comment{sampling $n_2$ indices with replacement from $\{1,\ldots,n\}$}
\State $\mathbf{Y}_2=\mathbf{Y}[ind]$; \Comment{resampled class labels}
\State $\mathbf{Z}_2=\mathbf{Z}[ind,:]$; \Comment{resampled encoder embedding}
\State $\mathbf{A}_2=\operatorname{zeros}(n_2,n_2)$; 
\For{$i=1,\ldots,n_2$}
\For{$j=i+1,\ldots,n_2$}
\State $\mathbf{A}_2[i,j]=\operatorname{Bernoulli}(\mathbf{Z}_2[i,\mathbf{Y}_2[j]])$;
\State $\mathbf{A}_2[j,i]=\mathbf{A}_2[j,i]$;
\EndFor
\EndFor
\State $pval=\operatorname{two sample}(\operatorname{GEE}(\mathbf{A},\mathbf{Y}), \operatorname{GEE}(\mathbf{A}_{2}, \mathbf{Y}_{2}))$;
\EndFunction
\end{algorithmic}
\end{algorithm}

\begin{figure}
\centering
\includegraphics[width=1.0\linewidth,trim={1cm 2cm 2cm 0.2cm},clip]{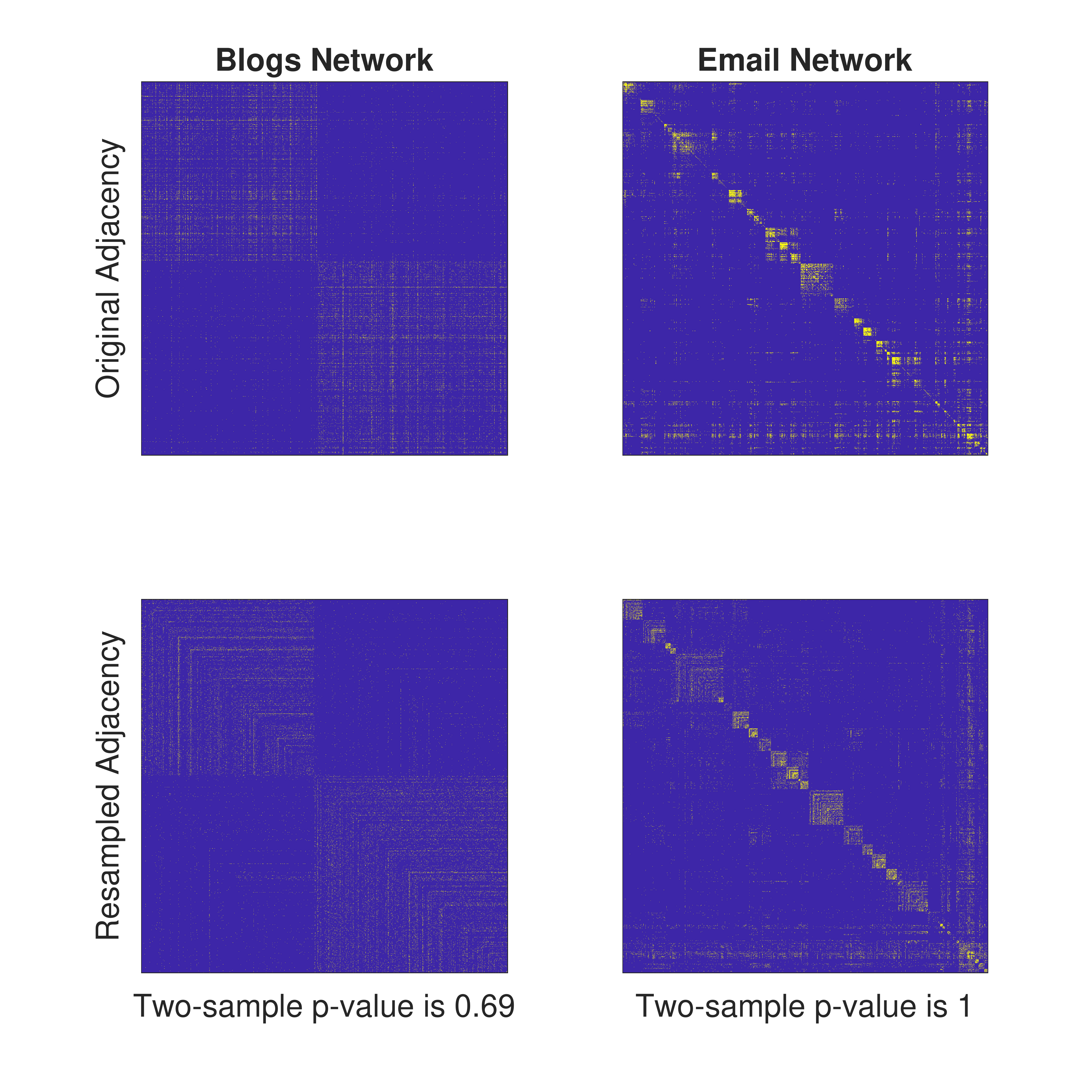}
\caption{The top row is the original adjacency matrix for each data, the bottom row is one bootstrap adjacency at $n=1000$, with two-sample test p-value computed at bottom.}
\label{fig3}
\end{figure}

Figure~\ref{fig3} visualizes the graph bootstrap results for the Political Blog and the Email Network (we pick these two graphs as they have clearer community structure and thus better for visualization). The resampled graph not only appears quantitatively similar to the original graph, but also yields a very large p-value from the two-sample test. This suggests the bootstrap graph is indiscernable from the original graph in distribution.

We repeat the bootstrap sampling at $n_2=1000$ for $1000$ times, and compute the two-sample p-value for each replicate. Algorithm~\ref{alg2} yields a mean p-value of $0.75$ for the political blog data. Moreover, only $0.4\%$ of the replicates yields a p-value that is less than $0.05$. In comparison, we also evaluated the naive bootstrap on graph adjacency. The mean p-value is $0.25$, and $26\%$ of the replicates have p-value less than $0.05$. Therefore, adjacency encoder embedding offers a better solution for graph bootstrap.

\section{Conclusion}
In this paper we proposed the one-hot graph encoder embedding method. The theoretical soundness is proved via asymptotic convergence and normality, and the numerical advantages are demonstrated in classification, clustering, and bootstrap. It is a flexible framework that can work with ground-truth labels, labels induced from other methods, partial or no labels at all. Most importantly, the excellent numerical performance is achieved via an elegant algorithmic design and a tiny fraction of time vs existing methods, making the graph encoder embedding very attractive and uniquely poised for huge graph data. 

\bibliographystyle{IEEEtran}
\bibliography{mgc,general}

\begin{IEEEbiography}
[{\includegraphics[width=1in,height=1.25in,clip,keepaspectratio]{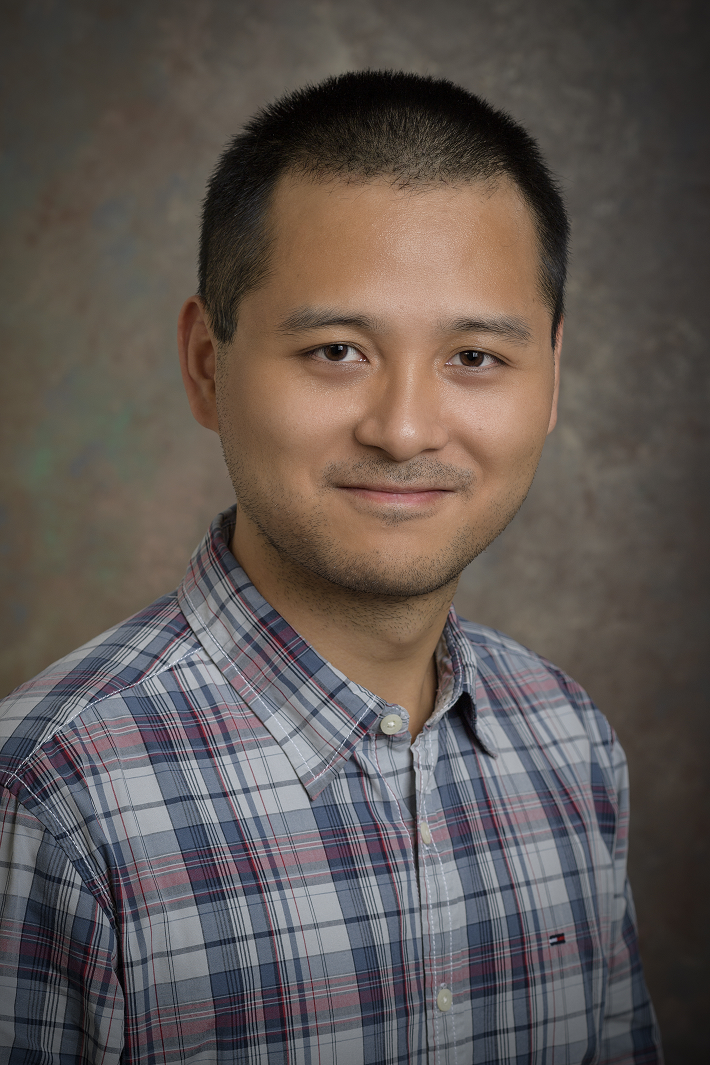}}]{Cencheng Shen received the BS degree in Quantitative Finance from National University of Singapore in 2010, and the PhD degree in Applied Mathematics and Statistics from Johns Hopkins University in 2015. He is 
assistant professor in the Department of Applied Economics and Statistics at University of Delaware. His research interests include graph inference, dimension reduction, hypothesis testing, correlation and dependence.}
\end{IEEEbiography}
\begin{IEEEbiography}
[{\includegraphics[width=1in,height=1.25in,clip,keepaspectratio]{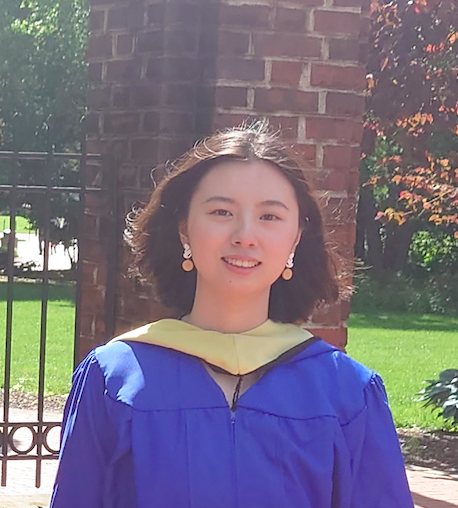}}]{Qizhe Wang received the BS degree in Statistics from University of Delaware in 2019, and the MS degree in Statistics from University of Delaware in 2021. }
\end{IEEEbiography}
\begin{IEEEbiography}
[{\includegraphics[width=1in,height=1.25in,clip,keepaspectratio]{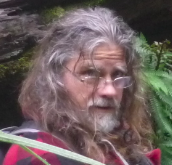}}]{Carey E. Priebe received the BS degree in mathematics from Purdue University in 1984, the MS degree in computer science from San Diego State University in 1988, and the PhD degree in information technology (computational statistics) from George Mason University in 1993. From 1985 to 1994 he worked as a mathematician and scientist in the US Navy research and development laboratory system. Since 1994 he has been a professor in the Department of Applied Mathematics and Statistics at Johns Hopkins University. His research interests include computational statistics, kernel and mixture estimates, statistical pattern recognition, model selection, and statistical inference for high-dimensional and graph data. He is a Senior Member of the IEEE, an Elected Member of the International Statistical Institute, a Fellow of the Institute of Mathematical Statistics, and a Fellow of the American Statistical Association.}
\end{IEEEbiography}

\if1\blind
\onecolumn
\setcounter{figure}{0}
\renewcommand{\thealgorithm}{C\arabic{algorithm}}
\renewcommand{\thefigure}{E\arabic{figure}}
\renewcommand{\thesubsection}{\thesection.\arabic{subsection}}
\renewcommand{\thesubsubsection}{\thesubsection.\arabic{subsubsection}}
\pagenumbering{arabic}
\renewcommand{\thepage}{\arabic{page}}

\bigskip
\begin{center}
{\large\bf APPENDIX}
\end{center}
\section{Proofs}
\label{sec:proofs}

Throughout the theorem proofs, without loss of generality we always assume:
\begin{itemize}
    \item $n_k =O(n)$;
    \item The class labels $\{Y_j,j=1,\ldots,n\}$ are fixed a priori.
\end{itemize}
The first assumption guarantees that each class is always non-trivial, which always holds when the class labels are being generated by a non-zero prior probability.

The second assumption assumes the class labels are fixed, or equivalently the proof is presented by conditioning on the class labels. This assumption facilitates the proof procedure and does not affect the results, because all asymptotic results hold regardless of the actual class labels, thus still true without conditioning. 

We shall prove Theorem~\ref{thmclt} and Corollary~\ref{thm1}:
\clt*

\one*

It is apparent that Corollary~\ref{thm1} is actually part of Theorem~\ref{thmclt}. In the proof, we shall start with proving the expectation and therefore proving the corollary first.

\subsection{Proof of Corollary~\ref{thm1}}
\begin{proof}
(i. SBM):\\
Under SBM, each dimension $k=1,\ldots,K$ of the vertex embedding satisfies
\begin{align*}
\mathbf{Z}_{i}[k] & = \mathbf{A}(i,:)\mathbf{W}(:,k) \\
                & = \frac{\sum_{j=1}^{n} I(Y_j=k)\mathbf{A}(i,j)}{n_k}\\
                & = \frac{\sum_{j=1, j\neq i, Y_j=k}^{n}  Bern(\mathbf{B}(y,Y_j))}{n_k}\\
                & = \frac{\sum_{j=1, j\neq i, Y_j=k}^{n} Bern(\mathbf{B}(y,k))}{n_k}.
\end{align*}
If $k=y$, the numerator is summation of $(n_k-1)$ i.i.d. Bernoulli random variables (since the summation includes a diagonal entry of $\mathbf{A}$, which is always $0$). Otherwise, we have $k \neq y$ and a summation of $n_k$ i.i.d. Bernoulli random variables. 

As $n$ increases, $n_k = O(n) \rightarrow \infty$ for any $k$. By law of large numbers it is immediate that
\begin{align*}
\mathbf{Z}_{i}[k] \rightarrow \mathbf{B}(y,k)
\end{align*}
dimension-wise. Concatenating every dimension of the embedding, we have
\begin{align*}
\|\mathbf{Z}_{i} - \mathbf{B}(y,:)\|_{2} \rightarrow 0.
\end{align*}

(ii. DC-SBM):\\
Under DC-SBM, we have
\begin{align*}
\mathbf{Z}_{i}[k] & = \mathbf{A}(i,:)\mathbf{W}(:,k) \\
                & = \frac{\sum_{j=1, j\neq i, Y_j=k}^{n} \mathbf{A}(i,j)}{n_k}\\
                & = \frac{\sum_{j=1, j\neq i, Y_j=k}^{n} \theta_i \theta_j Bern(\mathbf{B}(y,k))}{n_k}.
\end{align*}
The numerator in second line is a summation of either $n_k$ or $(n_k-1)$ independent Bernoulli random variables. Without loss of generality, we shall assume $k\neq y$ and $n_k$ summands in the third line, which is asymptotically equivalent. 

Note that given vertex $i$, $\theta_i$ and $\mathbf{B}(k,y)$ are fixed constants; and unlike the case of SBM, each random variable is now weighted by the degree parameter and thus not identical. Since all degree parameters lie in $(0,M]$, we have
\begin{align*}
Var(\mathbf{Z}_{i}[k]) &\leq \frac{\theta_i^2 M^2 \sum_{j=1, j\neq i, Y_j=k}^{n} \mathbf{B}(y,k)(1-\mathbf{B}(y,k))}{n_k^2} \leq \frac{\theta_i^2 M^2}{4n_k}.
\end{align*}
As $n \rightarrow \infty$ and $n_k =O(n)$, $Var(\mathbf{Z}_{i}[k]) \rightarrow 0$ for each dimension $k$.

By Chebychev inequality, $\mathbf{Z}_{i}[k]$ shall converge to its mean, which equals
\begin{align*}
E(\mathbf{Z}_{i}[k]) & = E(\frac{\sum_{j=1, j\neq i, Y_j=k}^{n} \theta_i \theta_j Bern(\mathbf{B}(y,k))}{n_k}) \\
& = \frac{\theta_i \sum_{j=1, j\neq i, Y_j=k}^{n} E(\theta_j) E(Bern(\mathbf{B}(y,k)))}{n_k} \\
& = \frac{\theta_i \mathbf{B}(y,k) \sum_{j=1, j\neq i, Y_j=k}^{n} E(\theta_j | Y_j=k)}{n_k} \\
&=\theta_i \mathbf{B}(y,k) \bar{\theta}_{(k)}^{(1)}.
\end{align*}
Concatenating every dimension of the embedding, it follows that
\begin{align*}
& \| \mathbf{Z}_{i} - \theta_i \mathbf{B}(y,:) \odot \bar{\Theta}^{(1)} \|_2 \rightarrow 0.
\end{align*}

(iii. RDPG):\\
Under RDPG, we have
\begin{align*}
\mathbf{Z}_{i}[k] & = \mathbf{A}(i,:)\mathbf{W}(:,k) \\
                & = \frac{\sum_{j=1, j\neq i, Y_j=k}^{n} \mathbf{A}(i,j)}{n_k}\\
                & = \frac{\sum_{j=1, j\neq i, Y_j=k}^{n} Bern(X_j^{T} x_i)}{n_k},
\end{align*}
and it suffices to assume $k\neq y$ and thus $n_k$ summands in the third line. Note that given vertex $i$ and its latent position, the randomness only comes from $X_j$ and Bernoulli.

The expectation satisfies 
\begin{align*}
E(\mathbf{Z}_{i}[k]) & = E( \frac{\sum_{j=1, j\neq i, Y_j=k}^{n} Bern(X_j^{T} x_i)}{n_k}) \\
& =  \frac{\sum_{j=1, j\neq i, Y_j=k}^{n} E(Bern(X_j^{T} x_i))}{n_k} \\
& = \frac{\sum_{j=1, j\neq i, Y_j=k}^{n} E(E(Bern(X_j^{T} x_i)|X_j))}{n_k} \\
& = \frac{\sum_{j=1, j\neq i, Y_j=k}^{n} E(X_j^{T} x_i)}{n_k} \\
& = \frac{ \sum_{j=1, j\neq i, Y_j=k}^{n} E(X_j^{T} x_i | Y_j=k) }{n_k} \\
& = \bar{\lambda}_{k}^{(1)}(x_i).
\end{align*}
And the variance satisfies
\begin{align*}
Var(\mathbf{Z}_{i}[k]) &= \sum_{j=1, j\neq i, Y_j=k}^{n} \frac{Var(Bern(X_j^{T} x_i))}{n_k^2}\\
&= \sum_{j=1, j\neq i, Y_j=k}^{n} \frac{Var(E(Bern(X_j^{T} x_i)|X_j))+E(Var(Bern(X_j^{T} x_i)|X_j))}{n_k^2}\\
&= \sum_{j=1, j\neq i, Y_j=k}^{n} \frac{Var(X_j^{T} x_i)+E((X_j^{T} x_i)(1-X_j^{T} x_i))}{n_k^2}\\
&= \sum_{j=1, j\neq i, Y_j=k}^{n} \frac{E((X_j^{T} x_i)^2)-E^2(X_j^{T} x_i)+E(X_j^{T} x_i)-E((X_j^{T} x_i)^2)}{n_k^2}\\
&= \sum_{j=1, j\neq i, Y_j=k}^{n} \frac{E(X_j^{T} x_i)- E^{2}(X_j^{T} x_i)}{n_k^2}\\
& \leq \frac{1}{4n_k}.
\end{align*}
As $n_k=O(n)$ and the numerator is bounded in $[0,\frac{1}{4}]$ (due to the valid probability constraint in RDPG), the variance converges to $0$ as sample size increases. Then by Chebychev inequality, $\mathbf{Z}_{i}[k] \rightarrow \bar{\lambda}_{k}^{(1)}(x_i)$. Concatenating every dimension of the embedding, it follows that \begin{align*}\|\mathbf{Z}_{i} - \bar{\lambda}_{x_i}^{(1)} \|_{2} \rightarrow 0.\end{align*}
\end{proof}

\subsection{Proof of Theorem~\ref{thmclt}}
\begin{proof}

(i. SBM):\\
From proof of Corollary~\ref{thm1} on SBM, it suffices to assume $k\neq y$ and
\begin{align*}
\mathbf{Z}_{i}[k] & = \frac{\sum_{j=1, j\neq i, Y_j=k}^{n} Bern(\mathbf{B}(y,k))}{n_k}.
\end{align*}
Applying central limit theorem to each dimension, we immediately have
\begin{align*}
\sqrt{n_k}(\mathbf{Z}_{i}[k] - \mathbf{B}(y,k)) \stackrel{d}{\rightarrow} \mathcal{N}(0,\mathbf{B}(y,k)(1-\mathbf{B}(y,k))). 
\end{align*}
Note that $\mathbf{Z}_{i}[k]$ and $\mathbf{Z}_{i}^{l}$ are always independent when $k \neq l$. This is because every vertex belongs to a unique class, so the same Bernoulli random variable never appears in another dimension. Concatenating every dimension yields
\begin{align*}
Diag(\vec{n})^{0.5} \cdot (\mathbf{Z}_{i} - \mathbf{B}(y,:)) \stackrel{d}{\rightarrow} \mathcal{N}(0,\Sigma_{\mathbf{B}_y}).
\end{align*}

(ii. DC-SBM):\\
From proof of Corollary~\ref{thm1} on DC-SBM, we have
\begin{align*}
\mathbf{Z}_{i}[k] & = \frac{\sum_{j=1, j\neq i, Y_j=k}^{n} \theta_i \theta_j Bern(\mathbf{B}(y,k))}{n_k}.
\end{align*}
Namely, each dimension of the encoder embedding is a summation of $n_k$ independent and weighted Bernoulli random variables that are no longer identical. 

Omitting the scalar constant $\theta_i$ in every summand, it suffices to check the Lyapunov condition for $\{U_j=\theta_j Bern(\mathbf{B}(k,y))\}$. Namely, prove that
\begin{align*}
\lim\limits_{n \rightarrow \infty}\frac{1}{s_{n_k}^{3}}\sum_{j=1}^{n_k}E(|U_j-E(U_j)|^3)=0
\end{align*}
where $s_{n_k}^2$ is the summation of variances of $\{U_j\}$. Based on the variance computation in the SBM proof, and note that all degree parameters $\theta_j$ are bounded in $(0,M]$, we have 
\begin{align*}
s_{n_k}^2 &= \sum_{j=1, j\neq i}^{n} Var(\theta_j|Y_j=k) I(Y_j=k) \mathbf{B}(y,k)(1-\mathbf{B}(y,k)) \\
&=O(n_k), \\
\sum_{j=1}^{n_k} E(|U_j-E(U_j)|^3) &= \sum_{j=1, j\neq i}^{n} E(\theta_j^3|Y_j=k) I(Y_j=k) E|Bern(\mathbf{B}(y,k))- \mathbf{B}(y,k)|^3 \\
&=O(n_k).
\end{align*}
It follows that
\begin{align*}
\frac{1}{s_{n_k}^{3}}\sum_{j=1}^{n_k}E(|U_j-E(U_j)|^3)=O(\frac{1}{\sqrt{n_k}}) \rightarrow 0,
\end{align*}
so the Lyapunov condition is satisfied.

Using Lyapunov central limit theorem and basic algebraic manipulation, we have
\begin{align*}
\sqrt{n_k}(\mathbf{Z}_{i}[k] - \theta_i \mathbf{B}(y,k) \bar{\theta}_{(k)}) \stackrel{d}{\rightarrow} \mathcal{N}(0,\theta_i^2 \bar{\theta}_{(k)}^{(2)} \mathbf{B}(y,k)(1-\mathbf{B}(y,k))). 
\end{align*}
Concatenating every dimension yields
\begin{align*}
Diag(\vec{n})^{0.5} \cdot (\mathbf{Z}_{i} - \theta_i \mathbf{B}(y,:) \odot \bar{\Theta}) \stackrel{d}{\rightarrow} \mathcal{N}(0, \ \theta_i^2 Diag(\bar{\Theta}^{(2)}) \cdot \Sigma_{\mathbf{B}_y}).
\end{align*}

(iii. RDPG):\\
From proof of Corollary~\ref{thm1} on RDPG, 
\begin{align*}
\mathbf{Z}_{i}[k] & = \frac{\sum_{j=1, j\neq i, Y_j=k}^{n} Bern(X_j^{T} x_i)}{n_k},
\end{align*}
which is a summation of $n_k$ independent Bernoulli random variables.

Next we check the Lyapunov condition for $\{U_j=Bern(X_j^{T} x_i)\}$:
\begin{align*}
s_{n_k}^2 &= \sum_{j=1, j\neq i}^{n} I(Y_j=k) Var(Bern(X_j^{T} x_i)) \\
&=O(n_k), \\
\sum_{j=1}^{n_k} E(|U_j-E(U_j)|^3) &= \sum_{j=1, j\neq i}^{n} I(Y_j=k) E|Bern(X_j^{T} x_i)- E(X_j^{T} x_i)|^3 \\
&=O(n_k).
\end{align*}
This is because the variance and the third moments are all bounded, due to the Bernoulli random variable and $X_j^{T} x_i$ being always bounded in $(0,1]$. It follows that
\begin{align*}
\frac{1}{s_{n_k}^{3}}\sum_{j=1}^{n_k}E(|U_j-E(U_j)|^3)=O(\frac{1}{\sqrt{n_k}}) \rightarrow 0,
\end{align*}
so the Lyapunov condition is satisfied.

Then from proof of Corollary~\ref{thm1} on RDPG variance, we have
\begin{align*}
Var(\mathbf{Z}_{i}[k]) &= \sum_{j=1, j\neq i, Y_j=k}^{n} \frac{E(X_j^{T} x_i)- E^{2}(X_j^{T} x_i)}{n_k^2}\\
&= \sum_{j=1, j\neq i, Y_j=k}^{n} \frac{E(X_j^{T} x_i | Y_j=k)- E^{2}(X_j^{T} x_i | Y_j=k)}{n_k^2}\\
& = \frac{\bar{\lambda}_{k}^{(1)}(x_i)-\bar{\lambda}_{k}^{(2)}(x_i)}{n_k}.
\end{align*}
By the Lyapunov central limit theorem, we have
\begin{align*}
\sqrt{n_k}(\mathbf{Z}_{i}[k] - \bar{\lambda}_{k}^{(1)}(x_i)) \stackrel{d}{\rightarrow} \mathcal{N}(0,\bar{\lambda}_{k}^{(1)}(x_i)-\bar{\lambda}_{k}^{(2)}(x_i)). 
\end{align*}
Concatenating every dimension yields
\begin{align*}
Diag(\vec{n})^{0.5} \cdot (\mathbf{Z}_{i} - \bar{\lambda}^{(1)}_{x_i}) \stackrel{d}{\rightarrow} \mathcal{N}(0, \ Diag(\bar{\lambda}^{(1)}_{x_i} - \bar{\lambda}^{(2)}_{x_i})).
\end{align*}
\end{proof}

\subsection{On Weighted Graph}
As mentioned in the main paper, the above theorems are readily applicable to weighted graph. For example, consider a weighted SBM:
\begin{align*}
\mathbf{A}(i,j) &\stackrel{i.i.d.}{\sim} U_{ij} \operatorname{Bernoulli}(\mathbf{B}(Y_i, Y_j))
\end{align*}
where $U_{ij} \sim U$ are independent and identically distributed bounded random variables. Then all the proof steps for SBM are intact, except the mean and variance in each dimension need to be multiplied by the mean and variance of the weight variable. Similarly for the DC-SBM and RDPG models. 

Therefore, so long the weight variable is independent and bounded, for the convergence it follows that
\begin{align*}
\|\mathbf{Z}_{i} - E(U) \mathbf{B}(y,:)\|_{2} \rightarrow 0;
\end{align*}
and for the normality it follows that
\begin{align*}
Diag(\vec{n})^{0.5} \cdot (\mathbf{Z}_{i} - E(U) \mu)  \stackrel{d}{\rightarrow}  \mathcal{N}(0,Var(U)\Sigma).
\end{align*}

\section{Simulation Details}

\subsection{Figure 2}
For each model in Figure 2, we always set $Y_i = 1,2$ with probability $0.5$ and $0.5$. The number of vertices is $n=2000$.
 
For the SBM graph, the block probability matrix is set to
\begin{align*}
\mathbf{B}=\begin{bmatrix}
0.13, 0.1\\
0.1, 0.13 
\end{bmatrix}.
\end{align*}

For DC-SBM, we set the block probability matrix as
\begin{align*}
\mathbf{B}=\begin{bmatrix}
0.9, 0.1 \\
0.1, 0.5
\end{bmatrix},
\end{align*}
then set $\theta_i \stackrel{i.i.d.}{\sim} Beta(1,4)$ for each $i$.

For RDPG, we generate the latent variable $X$ via the Beta mixture:
\begin{align*}
X_i \stackrel{i.i.d.}{\sim} \begin{cases}
Beta(1,5) \ \ \mbox{if $Y_i=1$;}\\
Beta(5,1) \ \ \mbox{if $Y_i=2$.}
\end{cases}
\end{align*}

\subsection{Figure 4 and Figure 5}
For each model, we always set $Y_i = 1,2,3$ with probability $0.2$, $0.3$, $0.5$ respectively. 
 
For the SBM graph, the block probability matrix is 
\begin{align*}
\mathbf{B}=\begin{bmatrix}
0.13, 0.1,0.1 \\
0.1, 0.13,0.1 \\
0.1, 0.1, 0.13
\end{bmatrix}.
\end{align*}

For DC-SBM, we set the block probability matrix as
\begin{align*}
\mathbf{B}=\begin{bmatrix}
0.9, 0.1, 0.1 \\
0.1, 0.5, 0.1 \\
0.1, 0.1, 0.2
\end{bmatrix},
\end{align*}
then set $\theta_i \stackrel{i.i.d.}{\sim} Beta(1,4)$ for each $i$. 

For RDPG, we generate the latent variable $X$ via the Beta mixture: 
\begin{align*}
X_i \stackrel{i.i.d.}{\sim} \begin{cases}
Beta(1,5) \ \ \mbox{if $Y_i=1$;}\\
Beta(5,5) \ \ \mbox{if $Y_i=2$;}\\
Beta(5,1) \ \ \mbox{if $Y_i=3$.}
\end{cases}
\end{align*}

\subsection{Figure E1}

We set $Y_i = 1,2$ with probability $0.5$ and $0.5$. Then for the SBM graphs in Figure E1, we set $n=3000$ and the block probability matrix as
\begin{align*}
\mathbf{B}=\begin{bmatrix}
0.2, 0.1\\
0.1, 0.1 
\end{bmatrix}
\end{align*}
for graph 1.

For graph 2, we set
\begin{align*}
\mathbf{B}=\begin{bmatrix}
0.1, 0.2\\
0.2, 0.1 
\end{bmatrix}.
\end{align*}

For graph 3, we set
\begin{align*}
\mathbf{B}=\begin{bmatrix}
0.1, 0.2\\
0.2, 0.4
\end{bmatrix}.
\end{align*}

For the DC-SBM models in Figure E1, we use the same block probability in each corresponding row, use $n=5000$, and generate $\theta_i \stackrel{i.i.d.}{\sim} Uniform(0.1,0.5)$.

For RDPG, we set $n=3000$ and generate the latent variable $X$ via the Beta mixture:
\begin{align*}
X_i \stackrel{i.i.d.}{\sim} \begin{cases}
Beta(2,3) \ \ \mbox{if $Y_i=1$;}\\
Beta(3,2) \ \ \mbox{if $Y_i=2$.}
\end{cases}
\end{align*}
for graph 1.

For graph 2, we let
\begin{align*}
X_i \stackrel{i.i.d.}{\sim} \begin{cases}
Uniform(0.15,0.25) \ \ \mbox{if $Y_i=1$;}\\
Uniform(0.1,0.2) \ \ \mbox{if $Y_i=2$.}
\end{cases}
\end{align*}

For graph 3, we let 
\begin{align*}
X_i \stackrel{i.i.d.}{\sim} \begin{cases}
Normal(0.15,0.01) \ \ \mbox{if $Y_i=1$;}\\
Normal(0.2,0.03) \ \ \mbox{if $Y_i=2$.}
\end{cases}
\end{align*}

\subsection{Normality Visualization}

\begin{figure}
\centering
\includegraphics[width=1.0\linewidth,trim={2cm 0cm 2cm -0.2cm},clip]{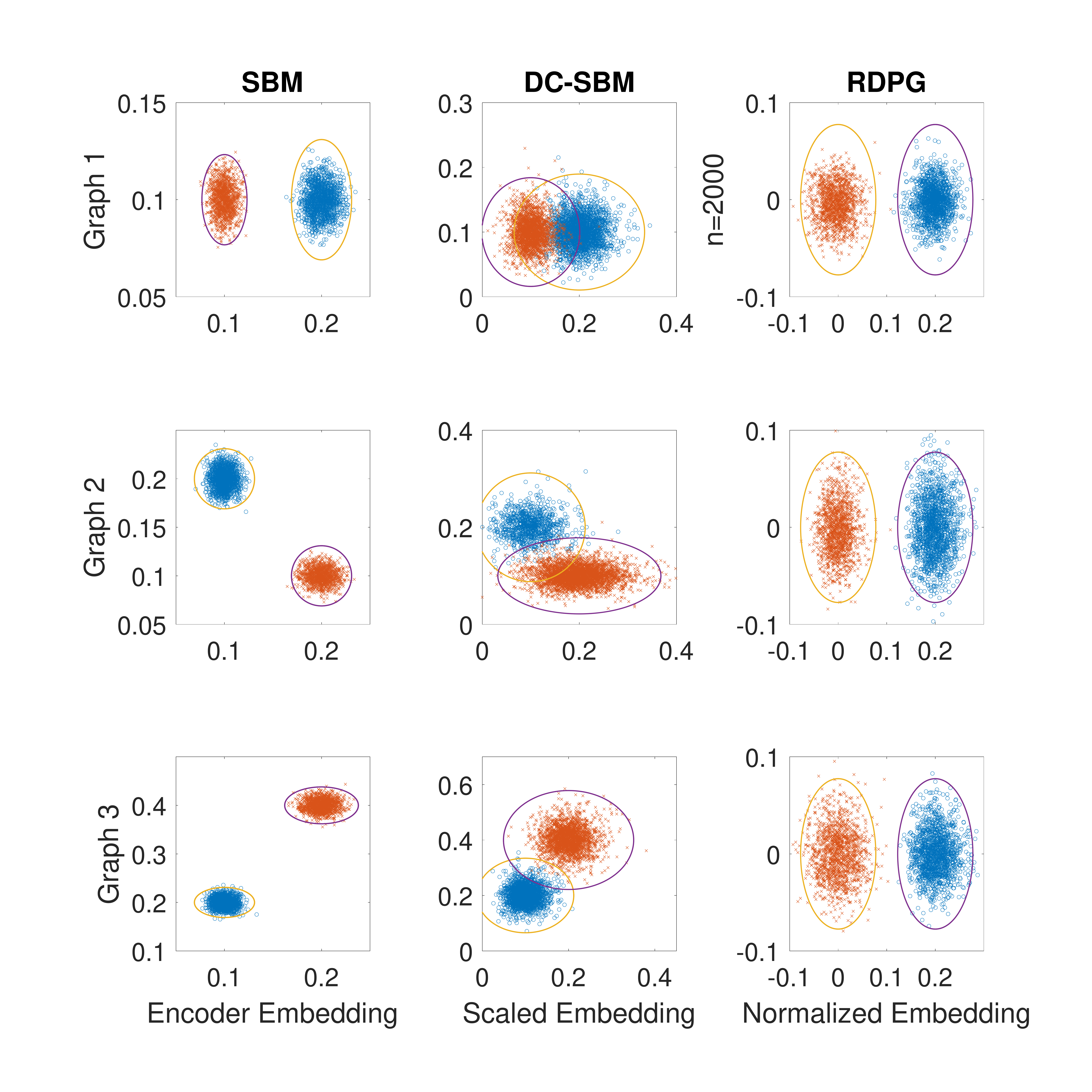}
\caption{Visualizing central limit theorem for the graph encoder embedding.}
\label{fig2}
\end{figure}

Figure~\ref{fig2} visualize the central limit theorem for the graph encoder embedding under SBM, DC-SBM, and RDPG graphs. To better visualize the normality and make sure each vertex has the same distribution, we plot the degree-scaled embedding for the DC-SBM graphs via $\mathbf{Z}_i / \theta_i ./ \bar{\Theta}$ (where $./$ denotes the entry-wise division); and plot a normalized encoder embedding for the RDPG graphs via normalizing $\mathbf{Z}_i$ by the mean and variance from Theorem~\ref{thmclt}, then add $0.2$ to all class $2$ vertices for clearer community separation. 

Then we draw two normality circles in every panel, using the class-conditional means as the center and three standard deviation from Theorem~\ref{thmclt} as the radius. Figure~\ref{fig2} clearly shows that the graph encoder embedding is approximately normally distributed in every case. 

\fi
\end{document}